%% file: lidarcap.tex
\documentclass[10pt,twocolumn,letterpaper]{article}

\usepackage{cvpr}              % To produce the CAMERA-READY version
\usepackage{graphicx}
\usepackage{amsmath}
\usepackage{amssymb}
\usepackage{booktabs}
\usepackage{float}
\usepackage{url}
\usepackage{soul}
\usepackage[accsupp]{axessibility}

\soulregister\cite7
\soulregister\cref7 
\newcommand{\myparagraph}[1]{\vspace{3pt}\noindent\textbf{#1}}

\usepackage[pagebackref,breaklinks,colorlinks]{hyperref}

\usepackage[capitalize]{cleveref}
\crefname{section}{Sec.}{Secs.}
\Crefname{section}{Section}{Sections}
\Crefname{table}{Table}{Tables}
\crefname{table}{Tab.}{Tabs.}

\def\cvprPaperID{11477} % *** Enter the CVPR Paper ID here
\def\confName{CVPR}
\def\confYear{2022}

\begin{document}

\title{LiDARCap: Long-range Marker-less 3D Human Motion Capture with LiDAR Point Clouds}

\author{Jialian Li\textsuperscript{1,}\thanks{Equal contribution.} \quad Jingyi Zhang\textsuperscript{1,}\footnotemark[1] \quad Zhiyong Wang\textsuperscript{1} \quad Siqi Shen\textsuperscript{1} \quad Chenglu Wen\textsuperscript{1}\quad Yuexin Ma\textsuperscript{2}  \\ Lan Xu\textsuperscript{2} \quad Jingyi Yu\textsuperscript{2}\quad Cheng Wang\textsuperscript{1,}\thanks{Corresponding author.}\\
\textsuperscript{1}Fujian Key Laboratory of Sensing and Computing for Smart Cities, Xiamen University \\
\textsuperscript{2}Shanghai Engineering Research Center of Intelligent Vision and Imaging, ShanghaiTech University\\
\tt\small \{szljl36,zhangjingyi1,wangzy\}@stu.xmu.edu.cn, \{siqishen,clwen,cwang\}@xmu.edu.cn, \\
\tt\small \{mayuexin,xulan1,yujingyi\}@shangtaitech.edu.cn
}

% \author{Jialian Li, Jingyi Zhang\\
% \and
\makeatletter
\let\@oldmaketitle\@maketitle% Store \@maketitle
\renewcommand{\@maketitle}{
    \@oldmaketitle% Update \@maketitle to insert...
    \begin{center}
        \vspace{-10mm}
        \includegraphics[width=\textwidth]{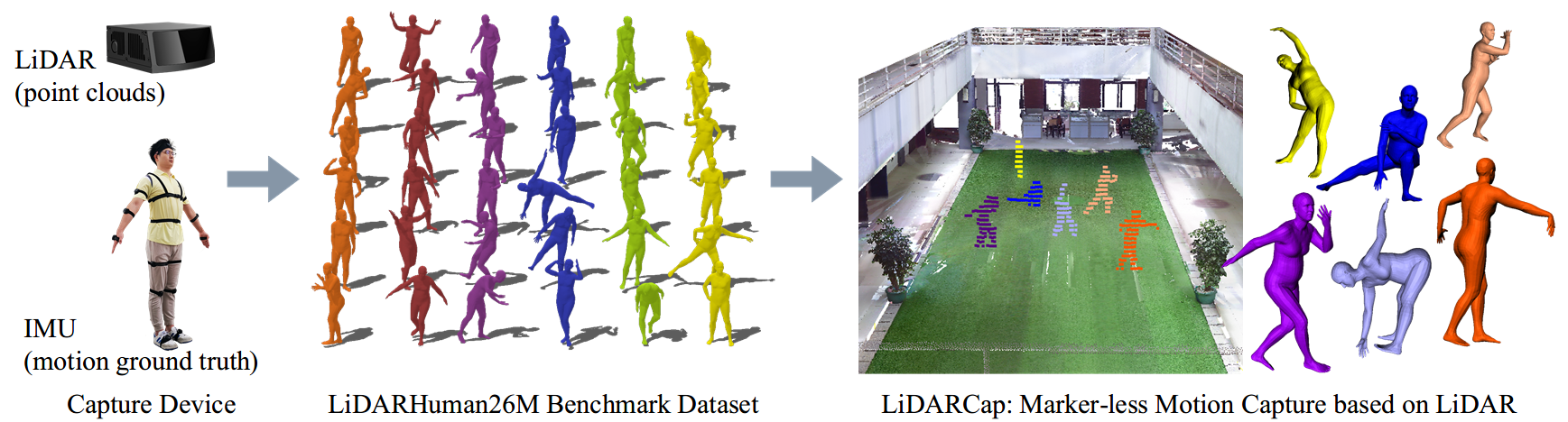}
    \end{center}
    \refstepcounter{figure}\normalfont Figure~\thefigure. Overview: The proposed LiDARHuman26M benchmark dataset consists of synchronous LiDAR point clouds, RGB images, and ground-truth  3D  human motions obtained from professional IMU devices, covering diverse motions and a large capture distance ranging. Based on LiDARHuman26M, we propose LiDARCap, a strong baseline motion capture approach on LiDAR point clouds, which achieves promising results as shown on the right end.
    \label{fig:overview}
    \newline
}
\makeatother
\maketitle

% \begin{figure*}
%   \centering
%   \fbox{\rule{0pt}{3in} \rule{0.9\linewidth}{0pt}}
%   %\includegraphics[width=0.8\linewidth]{egfigure.eps}

%   \caption{teaser}
%   \label{fig:onecol}
% \end{figure*}

%%%%%%%%% BODY TEXT

\input{section/abstarct}
\input{section/intro}
\input{section/related_work}
\input{section/approach}
\input{section/experiments}

\input{section/discussion}
\newpage

%%%%%%%%% REFERENCES
{\small
    \bibliographystyle{ieee_fullname}
    \bibliography{egbib}
}

\end{document}

%% file: section/abstarct.tex
%%%%%%%%% ABSTRACT
\begin{abstract}
\vspace{-3mm}
Existing motion capture datasets are largely short-range and cannot yet fit the need of long-range applications. We propose LiDARHuman26M, a new human motion capture dataset captured by LiDAR at a much longer range to overcome this limitation. Our dataset also includes the ground truth human motions acquired by the IMU system and the synchronous RGB images. We further present a strong baseline method, LiDARCap, for LiDAR point cloud human motion capture. Specifically, we first utilize PointNet++ to encode features of points and then employ the inverse kinematics solver and SMPL optimizer to regress the pose through aggregating the temporally encoded features hierarchically. Quantitative and qualitative experiments show that our method outperforms the techniques based only on RGB images. Ablation experiments demonstrate that our dataset is challenging and worthy of further research. Finally, the experiments on the KITTI Dataset and the Waymo Open Dataset show that our method can be generalized to different LiDAR sensor settings.
\end{abstract}
\vspace{-6mm}

%% file: section/intro.tex
\section{Introduction}
% 1. Marker-less mocap is important
The past ten years have witnessed a rapid development of marker-less human motion capture	~\cite{Davison2001,hasler2009markerless,StollHGST2011, wang2017outdoor}, with various applications like VR/AR and interactive entertainment.
However, conveniently capturing long-range 3D human motions in a large space remains challenging, which is critical for sports and human behavior analysis.

% 2. Issues of existing methods
So far, vision-based mocap solutions take the majority in this topic.
% 2.1 high-end RGB
The high-end solutions require dense optical markers~\cite{VICON,zhang2021we} or dense camera rigs~\cite{StollHGST2011,joo2015panoptic,collet2015high,TotalCapture} for faithfully motion capture, which are infeasible for consumer-level usage. 
% 2.2 monocular RGB --> degraded RGB
In contrast, monocular capture methods are more practical and attractive.
The recent learning-based techniques have enable robust human motion capture from a single RGB stream, using pre-scanned human templates~\cite{LiveCap2019tog,MonoPerfCap,EventCap_CVPR2020,DeepCap_CVPR2020,challencap} or parametric human models~\cite{OpenPose,Mehta2017,HMR18,AMASS_ICCV2019,SPIN_ICCV2019,VIBE_CVPR2020,SMPL2015}.
However, in the long-range capture scenarios where the performers are far away from the cameras, the captured images suffer from degraded and blurred artifacts, leading to fragile motion capture.
Various methods~\cite{xu20213d,xu20203d} explore to capture 3D human motions under such degraded and low-resolution images.
But such approaches are still fragile to capture the global positions under the long-range setting, especially when handling the textureless clothes or environment lighting changes.
% 2.3 IMU->based
In contrast, motion capture using body-worn sensor like Inertial Measurement Units (IMUs)~\cite{pons2011outdoor,huang2018DIP,Yi2021TransPoseR3} is widely adopted due to its environment-independent property.
However, the requirement of body-worn sensors makes them unsuitable to capture motions of people wearing everyday apparel.
Moreover, the IMU-based methods will suffer from an accumulated global drifting artifact, especially for the long-range setting.
% 2.3 commerical RGBD sensors
Those motion capture methods~\cite{dou-siggraph2016,Wei:2012,guo2017real,Su2020RobustFusionHV} using consumer-level RGBD sensors are also infeasible for the long-range capture in a large scene, due to the relatively short effective range (less than 5 m) of RGBD cameras.   

% Lidar
In this paper, we propose a rescue to the above problems by using a consumer-level LiDAR.
% its benefit
A LiDAR sensor provides accurate depth information of a large-scale scene with a large effective range (up to 30 m).
These properties potentially allow capturing human motions under the long-range setting in general lighting conditions, without suffering from the degraded artifacts of visual sensors.
% its challenges
Nevertheless, capturing long-range 3D human motions using a single LiDAR is challenging. 
First, under the long-range setting, the valid observed point clouds corresponding to the target performer is sparse and noisy,
making it difficult for robust motion capture.
Second, despite the popularity of LiDAR for 3D modeling, most existing work~\cite{shi2019pointrcnn,zhou2018voxelnet,lang2019pointpillars,milioto2019rangenet,hu2020randla,thomas2019kpconv} focus on scene understanding and 3D perception. 
The lack of a large-scale LiDAR-based dataset with accurate 3D human motion annotations leads to the feasibility of a data-driven motion capture pipeline using LiDAR.

% 3. Our key novelty
To tackle these challenges, we propose \textit{LiDARCap} -- the first marker-less, long-range and data-driven motion capture method using a single LiDAR sensor as illustrated in \cref{fig:overview}.
% 4. Our technical pipeline
% 4.1 dataset: 
More specifically, we first introduce a large benchmark dataset \textit{LiDARHuman26M} for LiDAR-based human motion capture.
Our dataset consists of various modalities, including synchronous LiDAR point clouds, RGB images and ground-truth 3D human motions obtained from professional IMU-based mocap devices~\cite{Noitom}.
It covers 20 kinds of daily motions and 13 performers with 184.0k capture frames, resulting in roughly 26 million valid  3D points of the observed performers with a large capture distance ranging from 12 m to 28 m.
Note that our LiDARHuman26M dataset is the first of its kind to open up the research direction for data-driven LiDAR-based human motion capture in the long-range setting. 
The multi-modality of our dataset also brings huge potential for future direction like multi-modal human behavior analysis.
% 4.2 baselines
Secondly, based on our novel LiDARHuman26M dataset, we provide LiDARCap, a strong baseline motion capture approach on LiDAR point clouds. 
% 4.3 evaluations
Finally, we provide a thorough evaluation of various stages in our LiDARCap as well as state-of-the-art image-based methods baselines using our dataset.
These evaluations highlight the benefit of the LiDAR-based method against the image-based method under the long-range setting.
We also provide preliminary results to indicate that LiDAR-based long-range motion capture remains to be a
challenging problem for future investigations of this new research direction. 
%5. Our technical contribution
To summarize, our main contributions include:
\begin{itemize}

	\item We propose the first monocular LiDAR-based approach for marker-less, long-range 3D human motion capture in a data-driven manner. 
	
	\item We propose a three-stage pipeline consisting of a temporal encoder, an inverse kinematics solver, and an SMPL optimizer to improve pose estimation performance.

	\item We provide the first large-scale benchmark dataset for LiDAR-based motion capture, with rich modalities and ground-truth annotations. The dataset will be made publicly available.

\end{itemize}

%% file: section/related_work.tex
% Related Work
\section{Related Work}
\label{sec:formatting}
%-------------------------------------------------------------------------
\myparagraph{Existing Pose Estimation Datasets.}
In recent years, deep networks have achieved impressive results in inferring the 3D human pose from images or video, and the research focus is tightly intertwined with dataset design. PennAction\cite{Zhang2013FromAT} and PoseTrack\cite{Andriluka2018PoseTrackAB} are the only ground-truth 2D video datasets, while InstaVariety\cite{Kanazawa2019Learning3H} and Kinetics-400\cite{Carreira2017QuoVA} are pseudo ground truth datasets annotated using a 2D keypoint detector. 
SURREAL\cite{Varol2017LearningFS} is a large-scale dataset with synthetically-generated but realistic images of people rendered from 3D sequences of human motion capture data. Those datasets have no 3D pose ground truth.

The Human3.6M\cite{Ionescu2014Human36MLS} dataset is a popular benchmark for pose estimation and captured in a controlled indoor environment. It has 3.6 million 3D human poses of 15 activities, and the 3D ground truth is collected using marker-based motion capture systems. Its goal is to predict the 3D locations of 32 joints in the human body defined by SMPL\cite{Bogo2016KeepIS}. HumanEva\cite{Sigal2009HumanEvaSV} is also restricted to indoor scenarios with static background, providing synchronized video with MoCap. MPI-INF-3DHP\cite{Mehta2017Monocular3H} is a multi-view dataset captured using a markerless motion capture system in a green screen studio, which records 8 actors performing 8 activities from 14 camera views. Meanwhile, it adopts foreground and background augmentation for addressing the scarcity and limited appearance variability. Another indoor dataset featuring synchronized video, marker-based ground-truth poses, and IMUs called TotalCapture\cite{Trumble2017TotalC3} labels for 1.9M frames. However, those datasets are all collected in indoor areas and have limited variability.

3DPW\cite{Marcard2018RecoveringA3} is an in-the-wild 3D dataset that captures SMPL body poses using IMU sensors and hand-held cameras. It contains 60 video sequences of several outdoor and indoor activities with 7 actors in 18 clothing styles, but 3DPW only provides images without depth information. The PedX\cite{Kim2019PedXBD} collects multi-modal pedestrians data at the large-scale outdoor scenario. Nevertheless, it only provides 3D pseudo label computed using the 2D annotations from a pair of stereo images and LiDAR point cloud. 
Based on the discussion and practical application requirements above, it is urgent to launch a dataset covering depth information and accurate 3D pose ground truth.

\myparagraph{Point Cloud Sequences Processing Methods.} Learning-based methods usually process point clouds by considering the Spatio-temporal relationship in point clouds along with time sequences. Choy et al.~\cite{Choy20194DSC} proposed 4D convolutional neural networks for the spatio-temporal perception that can directly process 3D-videos using high-dimensional convolutions.
Huang et al.~\cite{Huang2021SpatiotemporalSR} introduced a spatio-temporal representation learning framework, capable of learning from unlabeled 3D point clouds in a self-supervised fashion.
LatticeNet~\cite{Rosu2021LatticeNetFS} embeds raw point clouds into a sparse permutohedral lattice.
Wang et al.~\cite{Wang2021Selfsupervised4S} proposed a self-supervised schema to learn 4D spatio-temporal features from dynamic point cloud by predicting the temporal order of sampled and shuffled point cloud clips.
P4Transformer~\cite{Fan2021Point4T} proposes a Point 4D Transformer to model raw point cloud videos, consisting of a point 4D convolution and a transformer.
PSTNet~\cite{Fan2021PSTNetPS} proposes a point spatio-temporal convolution to achieve informative representations of point cloud sequences.
Wang et al.~\cite{Wang2021AnchorBasedSA} proposed anchor-based spatio-temporal attention 3D convolution operations to process dynamic 3D point cloud sequences.

\myparagraph{Pose Estimation Methods.} As an alternative to the widely used marker-based solutions~\cite{Vlasic2007PracticalMC,VMS, XTBV}, markerless motion capture~\cite{Bregler1998TrackingPW,Aguiar2008PerformanceCF} technologies alleviate the requirement of body-worn markers and have been widely investigated.
EventCap~\cite{Xu2020EventCapM3} combines model-based optimization and CNN-based human pose detection to capture high-frequency motion details and reduce the drifting in the tracking.
Li et al.~\cite{Li2020MonocularRV} proposed an approach to volumetric performance capture and novel-view rendering at real-time speed from monocular videos, eliminating the need for expensive multi-view systems or cumbersome pre-acquisition of a personalized template model.
RobustFusion~\cite{Su2020RobustFusionHV} proposes a human performance capture system combined with various data-driven visual cues using a single RGBD camera.
TailorNet~\cite{Patel2020TailorNetPC} proposes a neural model which predicts clothing deformation in 3D as a function of three factors: pose, shape, and style (garment geometry) while retaining wrinkle detail.
Zanfir et al.~\cite{Zanfir2021NeuralDF} presented a deep neural network to reconstruct people's 3D pose and shape from an RGB image, including hand gestures and facial expressions.
Given a single image and/or a single LiDAR sweep as input, S3\cite{yang2021s3} infers shape, skeleton and skinning jointly. However, they focus on the fusion of image and point cloud for human modeling and mainly rely on the synthetic dataset which is composed of the pedestrian behavior, such as walking and running in a short distance.

%% file: section/approach.tex
\section{Approach}
\subsection{Preliminaries}
Marker-less 3D motion capture in long-range scenarios is still challenging to the existing methods. As 2D cameras have no depth information, the inherent ambiguity of human joint locations exists in image-based methods, while depth cameras only work in near range. 
LiDAR sensors have the advantages of both long working range and good distinguish-ability in the depth dimension. In this work, we first develop a human motion dataset containing the LiDAR point clouds on long-range human motion scenarios, together with the synchronized IMU-captured motion ground truth. Our second goal is to establish an end-to-end model that can infer an optimal parametric human model from LiDAR point clouds. We use Skinned Multi-Person Linear Model(SMPL)\cite{Bogo2016KeepIS} to represent the pose and shape of a human body compactly. SMPL model contains pose parameters $\boldsymbol{\theta} \in \mathbb{R}^{72}$ associated with human motion, formulated as the relative rotations for 23 joints, to their parent joints and the global body rotation for the root joint, and the shape parameters $\boldsymbol{\beta} \in \mathbb{R}^{10}$, which control height, weight, limb proportions. The translation parameters $\mathbf{t} \in \mathbb{R}^{3}$ will be used when the human position is needed. The SMPL model deforms a template triangulated mesh with 6890 vertices based on pose and shape parameters, which is formulated as $\mathbf{V}=\mathcal{M}(\boldsymbol{\theta}, \boldsymbol{\beta})$. 

\subsection{Dataset: LiDARHuman26M}
Long-range motion capture has great potentials in various applications, such as immersive VR/AR experience and action quality assessment. In this paper, we propose the first long-range LiDAR-based motion capture dataset, LiDARHuman26M. 

\begin{figure}[t]
    \centering 
    %\fbox{\rule{0pt}{3in} \rule{0.9\linewidth}{0pt}}
    \includegraphics[width=\linewidth]{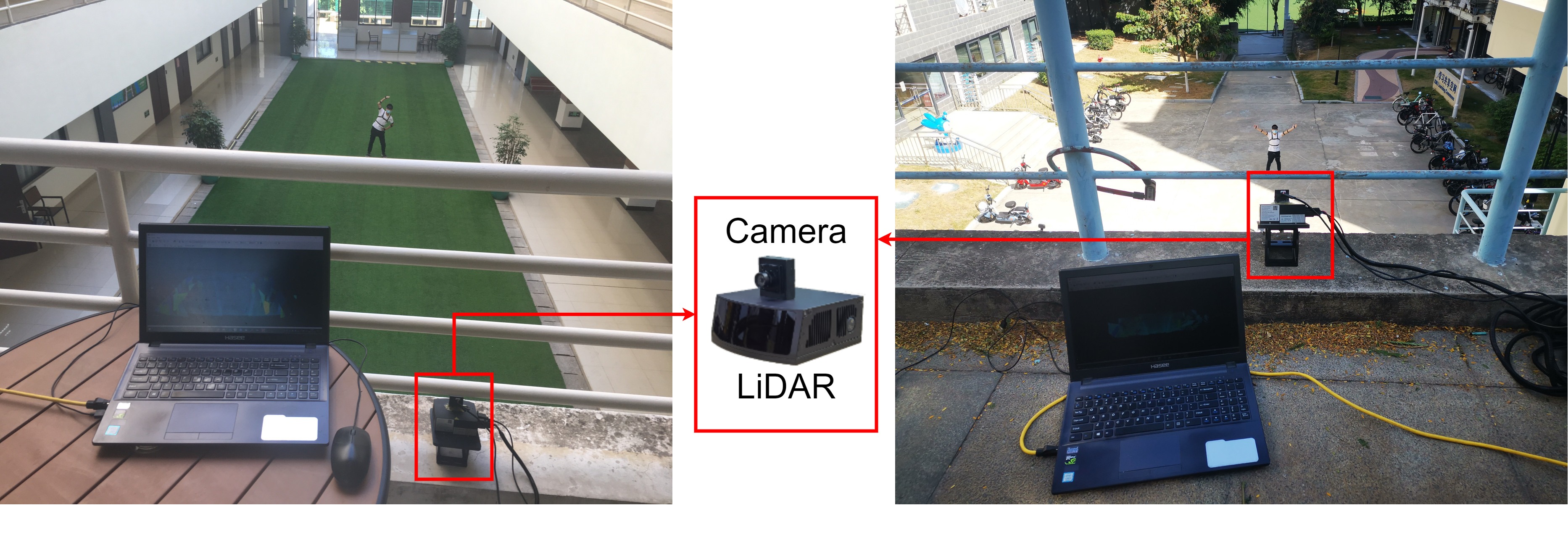}
    \vspace{-10mm}
    \caption{Two scenes for data acquisition.}
    \label{fig:data_acquisition}
    \vspace{-6mm}
\end{figure}

\myparagraph{Data Acquisition.} We collect data respectively in two scenarios as shown in \cref{fig:data_acquisition}. The first scene is a patio, which supports far distance human capture. The second scene is an open space between two buildings, supporting a large capturing pitch angle to avoid self-occlusion. The setup details of collection equipment are shown in  \cref{tab:setup details}.

\begin{table}
    \centering
    \resizebox{0.7\columnwidth}{!}{
        \begin{tabular}{l|cccccccc}
        \hline
        Scene &
        Range &
        Height & \\ 
        \hline
        The scene 1  & 12-28m    & 5m  \\
        The scene 2  & 14-24m    & 7m  \\
        \hline
        \end{tabular}
    }
    \vspace{-3mm}
    \caption{The setup details of equipment used in two scenes.}
    \vspace{-3mm}
    \label{tab:setup details}
\end{table}

We recruit 13 volunteers (including 11 males and 2 females) to participate in data collection, and they all have signed the consent. The duration for each one varies from 15 to 30 minutes. The distance distribution is shown in \cref{tab:dist_dist} In summary, LiDARHuman26M provides 184,048 frames, 26,414,383 points, and 20 kinds of daily motions (including walking, swimming, running, phoning, bowing, etc). It consists of three modalities: synchronous LiDAR point clouds, RGB images, and ground-truth 3D human motions from professional IMU-based mocap devices. We preprocessed the data by erasing the background and eliminating the localization error of the IMUs. Details are given in the supplementary materials.

\begin{table}
    \centering
    \resizebox{\columnwidth}{!}{
        \begin{tabular}{l|cccccc}
            \hline
            Dist(m)   & 11-13 & 14-16 & 17-19 & 20-22 & 23-25 & 26-28 \\
            \hline
            Ratio(\%) & 0.7   & 31.4  & 47.2  & 17.4  & 2.4   & 0.9 \\
            \hline
        \end{tabular}}
    \vspace{-3mm}
    \caption{Distance distribution in the dataset.}
    \vspace{-3mm}
    \label{tab:dist_dist}
\end{table}

\begin{table}
    \centering
    \resizebox{\columnwidth}{!}{
        \begin{tabular}{l|cccccccc}
\hline
  Dataset &
  Frames &
  Data Source &
  Long-range? &
  IMU? &
  Video? &
  Real? &
  Scene \\ \hline
Human3.6M\cite{Ionescu2014Human36MLS} & 3.6M & Image       & N & Y & Y & Y & Indoor  \\
HumanEva\cite{Sigal2009HumanEvaSV}  & 80.0K    & Image       & N & Y & N & Y & Indoor  \\
3DPW\cite{Marcard2018RecoveringA3}      & 51.0K    & Image       & N & Y & Y & Y & Outdoor \\
SURREAL\cite{Varol2017LearningFS}   & 6.5M & Image       & N & N & Y & N & Indoor  \\
PedX\cite{Kim2019PedXBD}      & 10.1K    & Point Cloud & Y & N & Y & Y & Outdoor \\
LiDARHuman26M  & 184.0K   & Point Cloud & Y & Y & Y & Y & Outdoor \\ \hline
\end{tabular}}
    \vspace{-3mm}
    \caption{Statistics and characteristics of related datasets.}
    \vspace{-3mm}
    \label{tab:other_datasets}
\end{table}
 
\myparagraph{Data Characteristic.} \cref{tab:other_datasets} presents statistics of our dataset in comparison to other publicly available 3D human pose datasets. 
{Our LiDARHuman26M dataset has the following features: First, our dataset contains many long-range (up to 28 meters away) human motions, while the image datasets usually have limited capturing distance. Although 3DPW has a certain improvement in this aspect, most of the annotated data still focuses on people nearby. Second, our dataset covers up to 20 daily motions, while HumanEva has only six motions and PedX mainly focuses on walking. Third, our dataset covers three different modalities, including point clouds, RGB videos, and the mocap ground truth provided by IMU. Current image-based datasets do not provide depth information, which is essential for long-range motion capture. SURREAL projects 3D SMPL meshes on the images, and the rendered images are unreal. PedX provides pseudo labels for 3D motions through optimization of LiDAR points along with 2D labels.}

\begin{figure}[t]
    \centering
    \includegraphics[width=\linewidth]{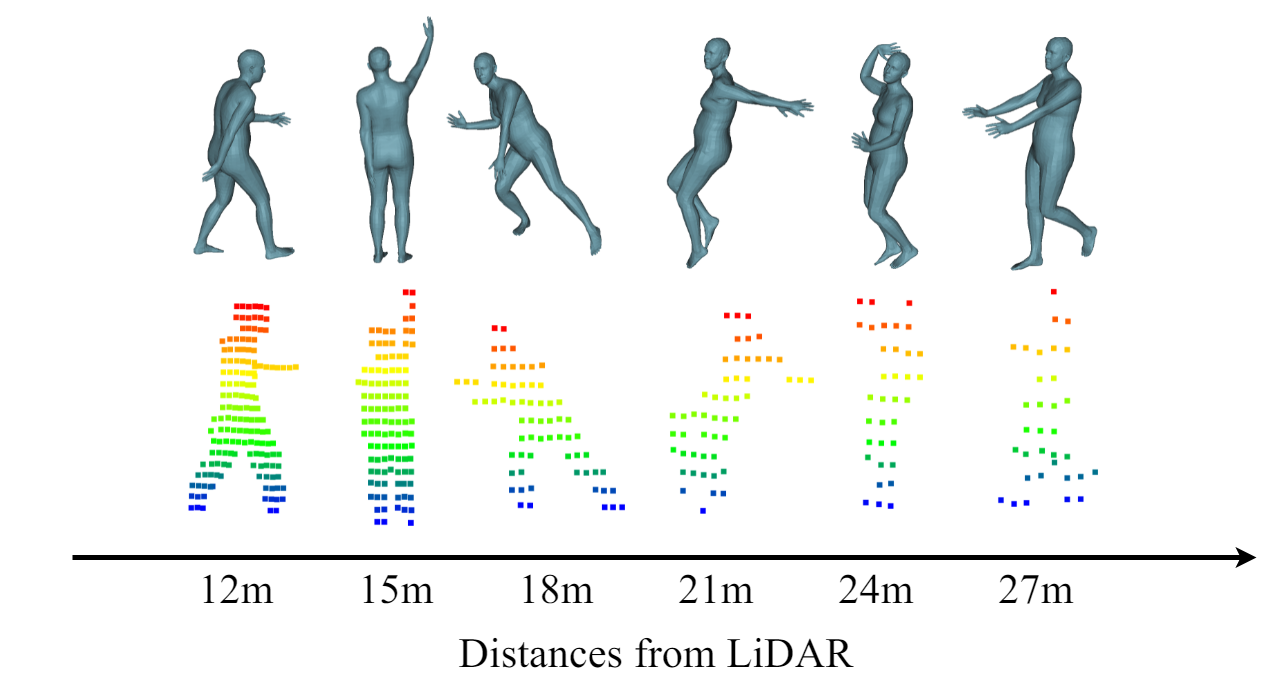}
    \vspace{-8mm}
    \caption{Different human poses with the distance to LiDAR increasing.}
    \vspace{-6mm}
    \label{fig:dataset}
\end{figure}

\myparagraph{Challenge.} The long-range characteristic of LiDARHuman26M causes sparsity. As shown in the \cref{fig:dataset}, the number of points on one person varies greatly, ranging from 30 points to 450 points. Furthermore, it can be manifested in whole body sparsity and partial missing. When the human body moves further from the LiDAR, the points that fall on the body are significantly reduced, resulting in insufficient information to describe the motion. Two different actions may have similar point cloud distributions at low resolution. For example, when capturing at 12m distance, the direction of the human head relative to the body is clear. The data ensures a good alignment between the captured motion and the rough outline. There are only one or two points on the human head at 24m and 27m capturing distance, which is insufficient to confirm the head orientation. 
Meanwhile, more parts of the body will inevitably miss with the distance increasing. For example, when capturing at 27m distance, the arm is missing, leading to a loss of elbow rotation. The possible reason behind this is body occlusion or too sparse points caused by too far capturing distance.

\begin{figure*}
    \centering
    \includegraphics[width=\linewidth]{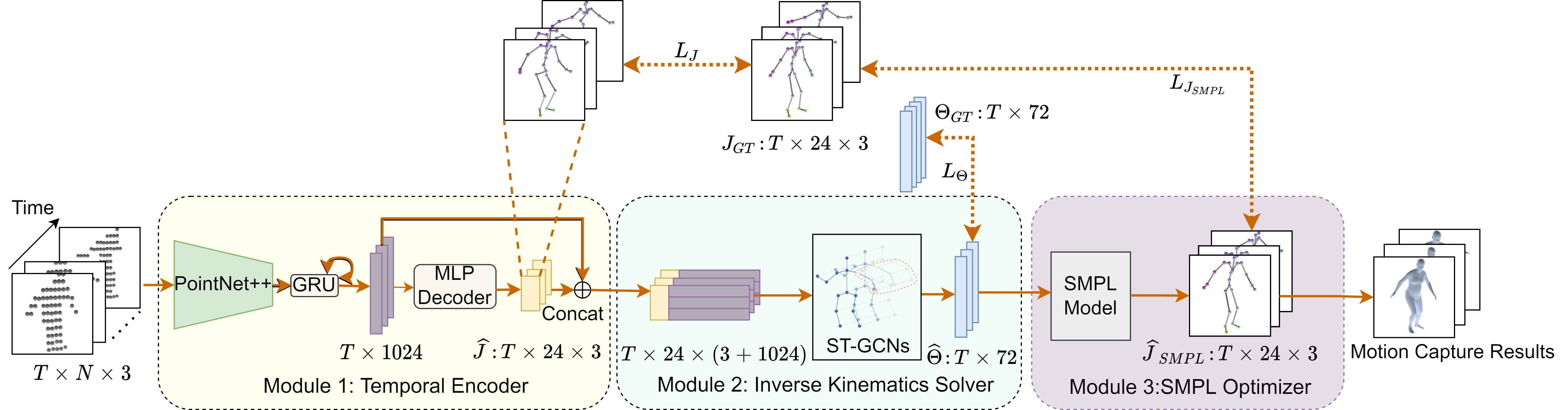}
    \vspace{-7mm}
    \caption{The pipeline of our method with a point cloud sequence as the input consists of a temporal encoder, an inverse kinematic solver, and an SMPL optimizer. $T$ represents the length of the sequence, and $N$ represents the number of points.}
    \vspace{-5mm}
    \label{fig:pipeline}
\end{figure*}

\subsection{Baseline: LiDARCap}
\label{sec:baseline}
We propose LiDARCap (shown in \cref{fig:pipeline}), a marker-less, long-range, and data-driven method for 3D human motion capture using LiDAR point clouds. Trained on LiDARHuman26M, LiDARCap takes point cloud sequences from monocular LiDAR sensor as input and outputs the 3D human motion sequences.

\myparagraph{Preprocessing.} Given an input LiDAR point cloud sequence
% $\mathbf{P}= \{\mathbf{p}_i \in \mathbb{R}^2| i = 1..N\}$
$\mathcal{P}  = \{\mathbf{P}^{(t)} | t=1...T\}$
of $T $ frames and each frame contains arbitrary number of points $\mathbf{P}^{(t)}=\{\mathbf{p}^{(t)}_i|i=1...n_t\}$. We fix the number to 512 by sampling or repeating to perform a unified down-sampling operations. 

\myparagraph{Temporal Encoder.} In this step, we leverage PointNet++\cite{qi2017pointnet} as the backbone to extract a 1024-dim global descriptor $\mathbf{f}^{(t)}$ for each point cloud frame $\mathbf{P}^{(t)}$.

In addition, in order to fuse temporal information, 
%the However, simply extracting point cloud features frame by frame is not enough to learn discriminative features. By adding the strong prior information of time, the extracted features can be fused between frames. The intuition behind this idea is that future motion capture can benefit from the motion information of the past frames. Similarly, we can add another direction to let the information from the future frames to the past frames. Formally,
the frame-wise features $\mathbf{f}^{(t)}$ are fed into a two-way GRU (biGRU) to generate hidden variables $\mathbf{g}^{(t)}$. At the last of this module, we use $\mathbf{g}^{(t)}$ as input to MLP decoders to predict the corresponding joint locations $\hat{\mathbf{J}}^{(t)} \in \mathbb{R}^{24\times3}$. Here, the loss $\mathcal{L}_{\mathcal{J}}$ of the temporal encoder is formulated as:
\begin{equation}
    \mathcal{L}_{\mathcal{J}} = \sum_t \Vert \mathbf{J}_{GT}^{(t)}- \hat{\mathbf{J}}^{(t)}\Vert_2^2
\end{equation}
where $\mathbf{J}_{GT}^{(t)}$ is the ground truth joint locations of the t-th frame.

\myparagraph{Inverse Kinematics Solver.} ST-GCN\cite{stgcn2018aaai} is adopted as the backbone here to extract features of the predicted joints in a graph way. We concatenate the frame-wise global feature with each joint to generate the completed joint features $\mathbf{Q}^{(t)} \in \mathbb{R}^{24\times(3+1024)}$ as the graph node. The output of ST-GCN is subsequently fed into the regressor to compute the joint rotations $\mathbf{R}^{(t)}_{6D} \in \mathbb{R}^{K\times6}$. The 6D rotation is mapped to the final axis-angle format when the loss is computed. We choose the 6D rotation representation as the intermediate results for its better continuity, as demonstrated in \cite{Zhou2019OnTC}.

The loss of this module $\mathcal{L}_{\boldsymbol{\Theta}}$ is formulated as:
\begin{equation}
    \mathcal{L}_{\boldsymbol{\Theta}} = \sum_t \Vert \boldsymbol{\theta}_{GT}^{(t)}- \hat{\boldsymbol{\theta}}^{(t)}\Vert_2^2
\end{equation}
where $\boldsymbol{\theta}_{GT}^{(t)}$ is the ground truth pose parameters of the t-th frame .

\myparagraph{SMPL Optimizer.} We put an SMPL Optimizer module at the last stage to further improve the regression on $\boldsymbol{\theta}$. The joint rotations are fed into an off-the-shelf SMPL model to obtain the 24 joints on the SMPL mesh. $\mathcal{L}_2$ loss between the predicted joints and the ground truth ones is used again in this module to increase the accuracy of the regressed $\boldsymbol{\theta}$ in the last stage. The only difference is that the joints in the first stage are regressed directly through an MLP-based decoder, and here the joints are sampled on the parametric mesh vertices determined by $\boldsymbol{\theta}$.

The loss of this module $\mathcal{L}_{\mathcal{J}_{SMPL}}$ is formulated as:
\begin{equation}
    \mathcal{L}_{\mathcal{J}_{SMPL}} = \sum_t \Vert {\mathbf{J}_{GT}^{(t)}}- \hat{\mathbf{J}}^{(t)}_{SMPL}\Vert_2^2
\end{equation}
where $\mathbf{J}^{(t)}_{SMPL}$ is the joint locations sampled from the SMPL mesh parameterized by the pose parameter $\hat{\boldsymbol{\theta}}^{(t)}$.

This step provides stronger constraints on the regression of $\boldsymbol{\theta}$ in a geometrically intuitive way. The ablation experiment is conducted to demonstrate its necessity, and more details can be seen in \cref{sec:evaluation}. 

To sum up, our pipeline can be trained through optimizing the united loss function $\mathcal{L}$ formulated as below in an end-to-end way:
\begin{equation}
    \mathcal{L} = \mathcal{L}_{\mathcal{J}} + \mathcal{L}_{\boldsymbol{\Theta}} +  \mathcal{L}_{\mathcal{J}_{SMPL}}
\end{equation}

\myparagraph{Training details.} We train our method for 200 epochs with Adam optimizer \cite{Kingma2015AdamAM} and set the dropout ratio as 0.5 for the GRU layers and ST-GCN module. We apply batch normalization layer after every convolutional layer except the final output layer before the decoder. During training, one NVIDIA GeForce RTX 3090 Graphics Card is utilized. The batch size is set to be 8, while the learning rate is set to be $1\times10^{-4}$. The decay rate is $1\times10^{-4}$. The network architecture involved in the evaluation section is trained using the most suitable learning rate until convergence. We train our method on the proposed LiDARHuman26M dataset, and experiment details are provided in \cref{sec:experiments}.
% \newpage
% []
% \newpage
% []
% \newpage

\begin{figure*}
    \centering
    \includegraphics[width=0.9\textwidth]{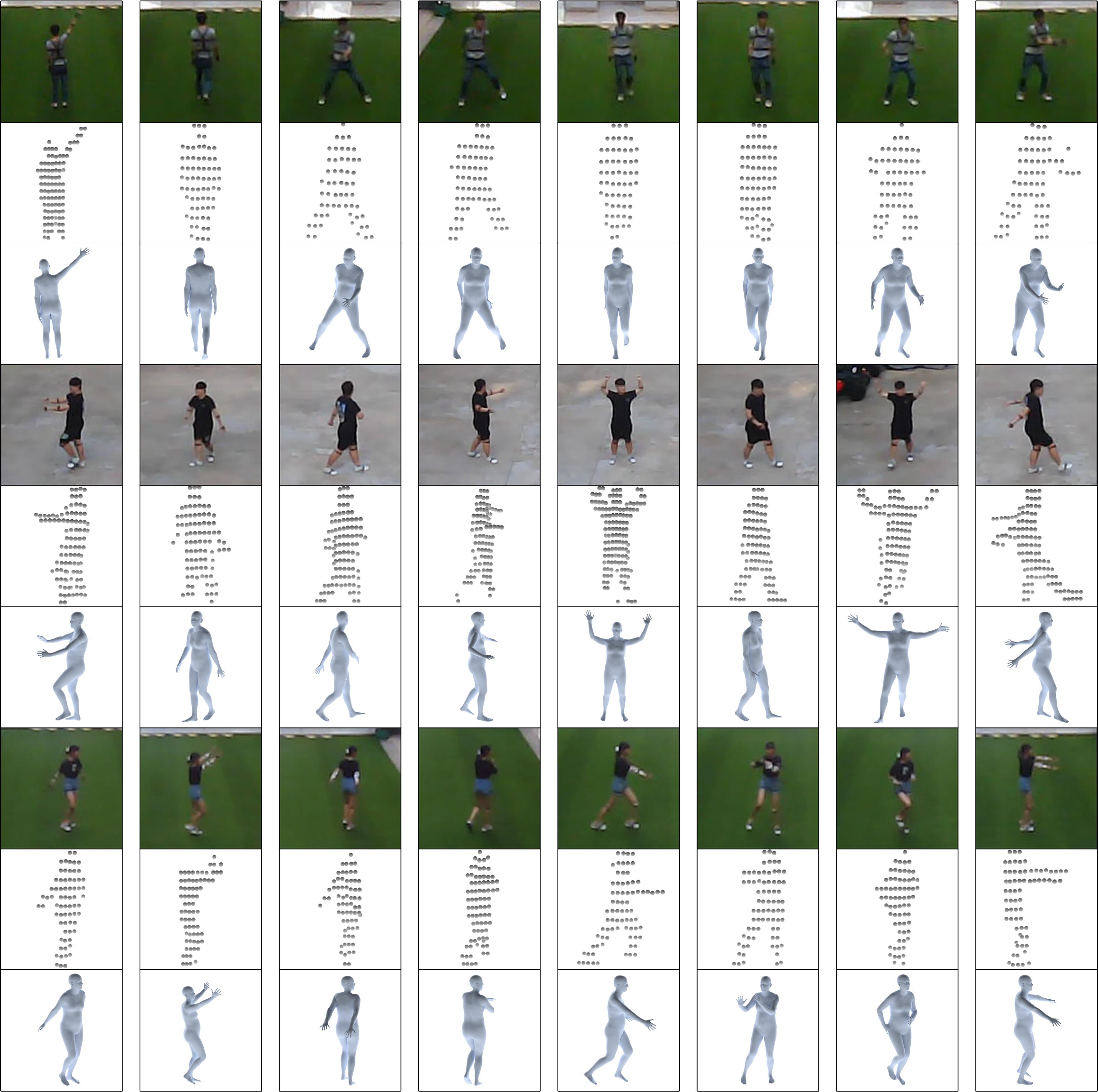}
    \vspace{-2mm}
    \caption{3D capturing results on long-range human motions. For each body motion, the top row shows the reference images, the middle row shows the input LiDAR points, and the bottom row shows the captured motion results on the LiDAR view.} % 我看challencap中没有关于results gallery的介绍
    \vspace{-6mm}
    \label{fig:results gallery}
\end{figure*}

%% file: section/experiments.tex
\section{Experiments}
\label{sec:experiments}
\subsection{Comparison} % 和HMR和VIBE的对比
The proposed LiDARCap method performs well in predicting human motions in long-range scenarios, as shown in \cref{fig:results gallery}. For further investigation, our method was compared with the state-of-the-art (SOTA) image-based motion capture methods. Quantitative and qualitative comparisons with HMR\cite{HMR18} and VIBE\cite{VIBE_CVPR2020} are conducted where the latter also relies on the temporal encoding. 

\begin{figure}[t]
    \centering
    \includegraphics[width=\linewidth]{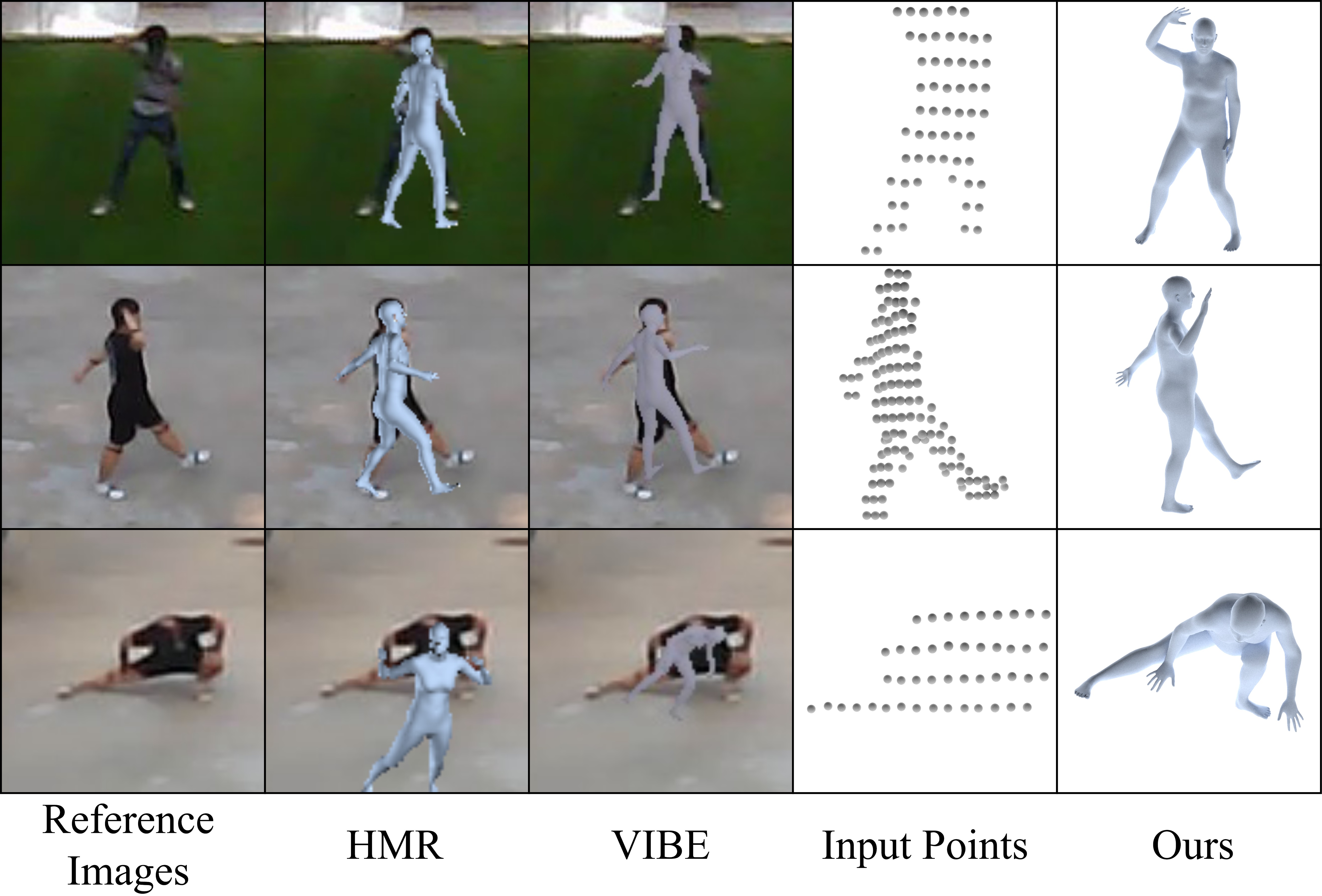}
    \vspace{-4mm}
    \caption{Qualitative comparison with the image-based methods. Our method provides accurate human pose, while the results of SOTA image-based methods contain large errors.}
    \vspace{-2mm}
    \label{fig:comparison}
\end{figure}

As shown in the \cref{fig:comparison}, benefited from the 3D distinguish-ability of the LiDAR point clouds, our method outperforms the image-based methods. The performance of HMR is badly contaminated by the low quality of the distant images, while VIBE can speculate some unclear motions with the help of sequential constrain.

\cref{tab:comparison} shows the corresponding quantitative comparisons using different evaluation metrics. We report Procrustes-Aligned Mean Per Joint
Position Error (PA-MPJPE), Mean Per Joint Position Error (MPJPE), Percentage of Correct Keypoints (PCK), and Per Vertex Error (PVE). The error metrics are measured in millimeters. In addition, Acceleration error$(m/s^2)$ is also recorded as an important evaluation indicator for sequence data. Benefiting from the effective use of 3D spatial information, our method significantly outperforms HMR and VIBE.

\subsection{Evaluation}
\label{sec:evaluation}
To study the effect of different components of our method, we conduct two ablation experiments. The first experiment validates the effectiveness of the combination of PointNet++ and ST-GCN. The second one verifies the effectiveness of the combination of the inverse kinematics solver and SMPL optimizer.

\myparagraph{Evaluation on network structure.} For simplicity, our method is called P++/ST-GCN. On the one hand, we replace the PointNet++(P++) backbone with other diligently-designed network structures. They are P4Transformer(P4T)\cite{Fan2021Point4T}, attention(ATT) module in \cite{Jiang2019SkeletonAware3H} and the voting(VOT) module in \cite{Li2019PointToPoseVB}. In order to fuse the spatial-temporal information, we use the P4Transformer instead of the original one as the backbone of the latter two. On the other hand, we need to evaluate whether it is necessary to leverage ST-GCN to exploit the joint features over the temporal dimension instead of biGRU. \cref{tab:network_architecture} shows the comparison results mentioned above, from which we find that on our dataset, the more complicated operation like attention and voting will result in a decrease in the performance. Global features help achieve the best performance, and there is no significant difference between P++ and P4T. Moreover, introducing the kinematic tree can help localize the adjacent joints better than the biGRU, which can only impact the frame-wise global features. The convolution on the same joints over the time step also ensures the continuity and consistency explicitly.

\begin{table}
    \centering
    \resizebox{\columnwidth}{!}{
        \begin{tabular}{l|cccccc}
            \hline
            Method & MPJPE$\downarrow$ & PA-MPJPE$\downarrow$ & PCK0.5$\uparrow$ & PCK0.3$\uparrow$ & Accel$\downarrow$ & PVE$\downarrow$ \\ \hline
            HMR & 224.86 & 130.71 & 0.67 & 0.49 & 22.07 & 284.15 \\
            VIBE & 154.61 &  108.19 & 0.82 & 0.64  & 12.49  & 191.55 \\
            Ours & \textbf{79.31} & \textbf{66.72} & \textbf{0.95} & \textbf{0.86} & \textbf{4.52} & \textbf{101.64}  \\ \hline
        \end{tabular}}
    \caption{Quantitative comparison of our method and image-based methods in terms of capturing accuracy.}
    \vspace{-2mm}
    \label{tab:comparison}
    \vspace{-2mm}
\end{table}

\begin{table}
    \centering
    \resizebox{\columnwidth}{!}{
        \begin{tabular}{l|cccccc}
            \hline
            Method & MPJPE$\downarrow$ & PA-MPJPE$\downarrow$ & PCK0.5$\uparrow$ & PCK0.3$\uparrow$ & Accel$\downarrow$ & PVE$\downarrow$ \\ \hline
            P++/GRU & 86.43 & 72.19 & 0.94 & 0.83 & 5.20 & 109.48 \\
            ATT/ST-GCN & 96.28 & 75.21 & 0.92 & 0.81 & 4.66 & 120.42 \\
            P4T/ST-GCN & 79.52 & \textbf{66.25} & \textbf{0.95} & \textbf{0.86} & 4.54 & 101.77 \\
            VOT/ST-GCN & 146.20 & 100.40 & 0.83 & 0.67 & 7.10 & 185.33 \\
            P++/ST-GCN(Ours)   & \textbf{79.31} & 66.72 & \textbf{0.95} & \textbf{0.86} & \textbf{4.52} & \textbf{101.64}  \\ \hline
        \end{tabular}}
    \caption{Quantitative evaluation of different encoders and sequential processing methods.}
    \vspace{-4mm}
    \label{tab:network_architecture}
\end{table}

\myparagraph{Evaluation on stages.} The necessity of all the three modules proposed in \cref{sec:baseline} is demonstrated in \cref{tab:stage}. Among the three modules, the temporal encoder used to extract point features is indispensable. The performance difference is that the joint locations help the network learn the motion features more efficiently both in the first and the third stage. Among them, the previous one is used to constrain the solution domain of the rotations, while the latter one serves an important role as a posteriori validation.

\begin{table}
    \centering
    \resizebox{\columnwidth}{!}{
        \begin{tabular}{l|cccccc}
            \hline
            Method & MPJPE$\downarrow$ & PA-MPJPE$\downarrow$ & PCK0.5$\uparrow$ & PCK0.3$\uparrow$ & Accel$\downarrow$ & PVE$\downarrow$ \\ \hline
            P++ w/o ST-GCN & 85.93 & 70.61 & 0.94 & 0.84 & 4.91 & 109.27 \\ P++/ST-GCN w/o $\mathcal{L}_{\mathcal{J}_{SMPL}}$ 
             & 87.13 & 69.35 & 0.94 & 0.83 & 4.98 & 110.23 \\
            P++/ST-GCN(Ours)   & \textbf{79.31} & \textbf{66.72} & \textbf{0.95} & \textbf{0.86} & \textbf{4.52} & \textbf{101.64}  \\ \hline
        \end{tabular}}
    \caption{Quantitative evaluation of different combinations of stages.}
    \vspace{-8mm}
    \label{tab:stage}
\end{table}

\subsection{Distance Analysis} 
\begin{figure}
  \centering
  \begin{subfigure}{\columnwidth}
    \includegraphics[width=\columnwidth]{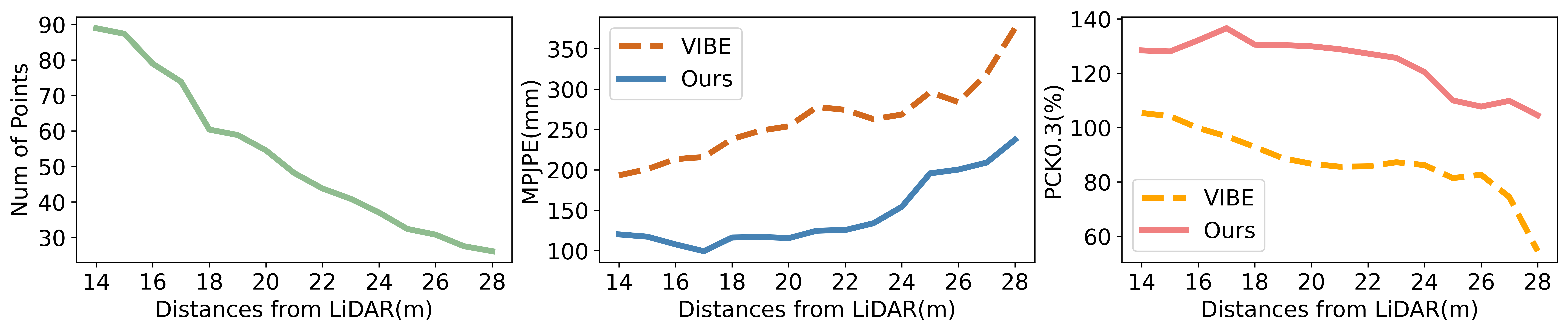}
    \caption{The number of points and performance of VIBE and ours over the distance.}
    \label{fig:dist_plot}
  \end{subfigure}

  \begin{subfigure}{\columnwidth}
    \includegraphics[width=\columnwidth]{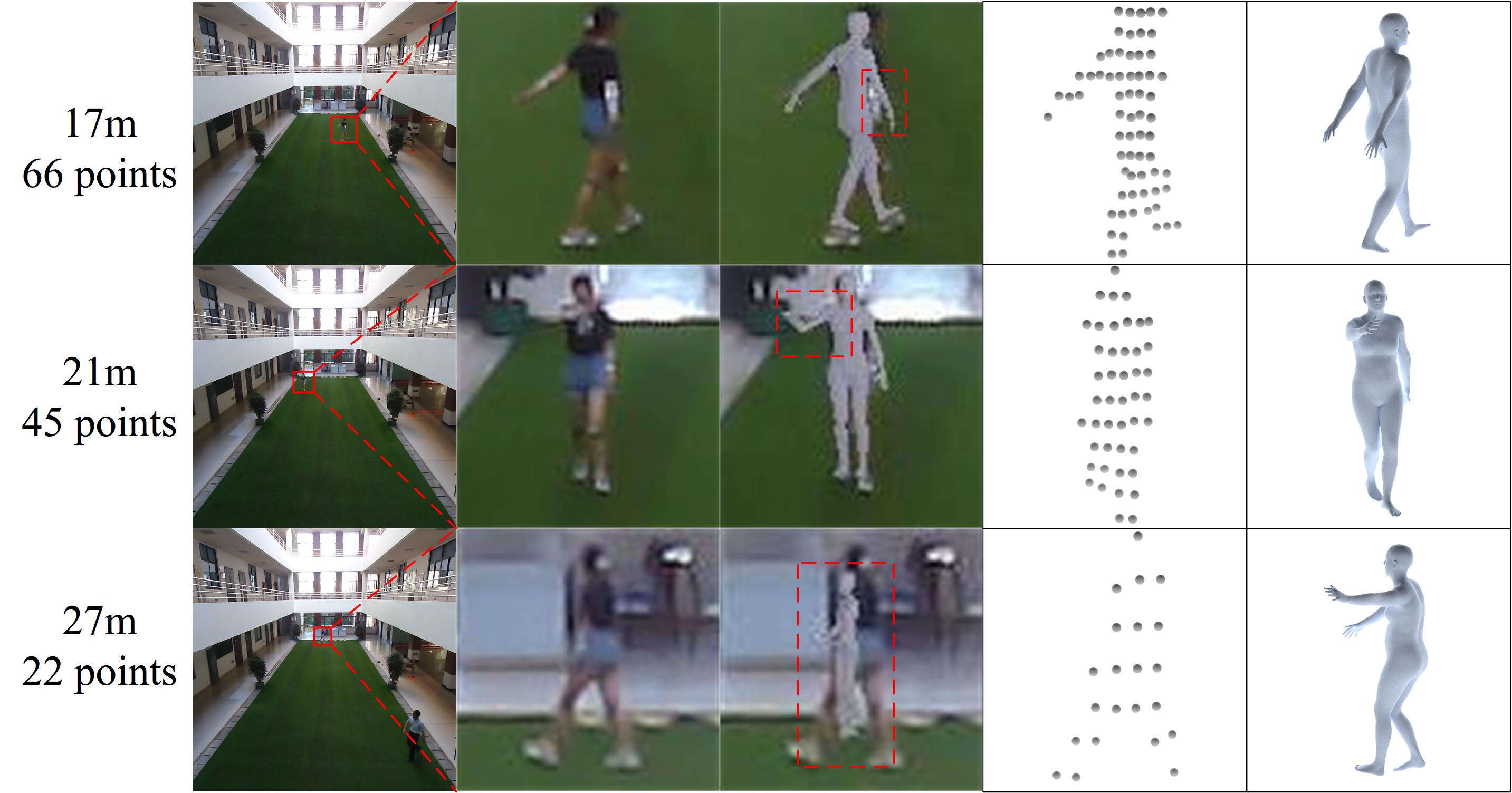}
    \caption{The qualitative results over the chosen three distances. The columns from left to right are the original images, the enlarged images, VIBE results, the point clouds, and our results.}
    \label{fig:dist_visu}
  \end{subfigure}
  \caption{Evaluation of VIBE and ours over the different distances. The figures show that performance will decrease as distance increases. However, our algorithm can still achieve more convincing results than VIBE. }
  \label{fig:short}
  \vspace{-5mm}
\end{figure}

Benefiting from the excellent characteristics of the point cloud, our method has achieved good results in long-range motion capture. However, the latest version of the algorithm still cannot do well at a too far distance. We take the sequence with the most dramatic distance change as the illustration, in which the trajectory of the volunteer ranges from 15 to 28 meters from the LiDAR. As can be seen from \cref{fig:dist_plot}, the number of points projected on a person decreases sharply as the distance increases, and at the farthest distance, the number is less than 30. At this moment, apart from the outline of the human, it is difficult to discriminate the detailed movements. The low resolution of long-range images also poses great challenges to image-based ones. Although the performance of our method and VIBE decreases as the distance increases, we can still maintain better results which is shown in \cref{fig:dist_visu}. This is because, with the distance increases, the human can still be segmented clearly from the background in the point cloud while the image pixels of human is more prone to mix with the background ones, though the resolution of both data sources is decreasing.

\begin{figure}[t]
    \centering
    \includegraphics[width=\linewidth]{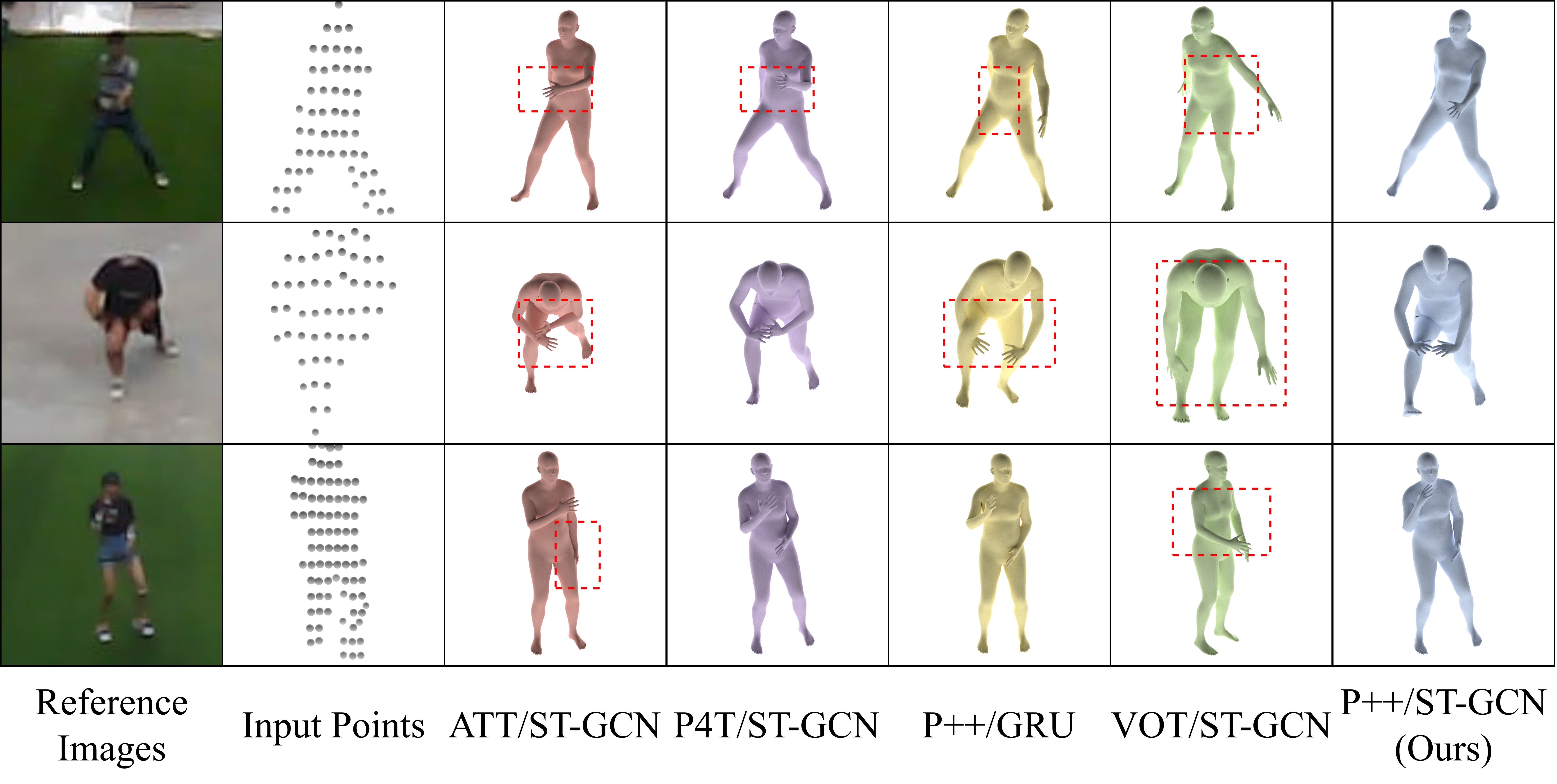}
    \vspace{-8mm}
    \caption{Qualitative results of different network structures. P4T/ST-GCN and our method can capture more consistent motions than other methods.}
    \vspace{-3mm}
    \label{fig:net_struct}
\end{figure}

\begin{figure}[t]
    \centering
    \includegraphics[width=\linewidth]{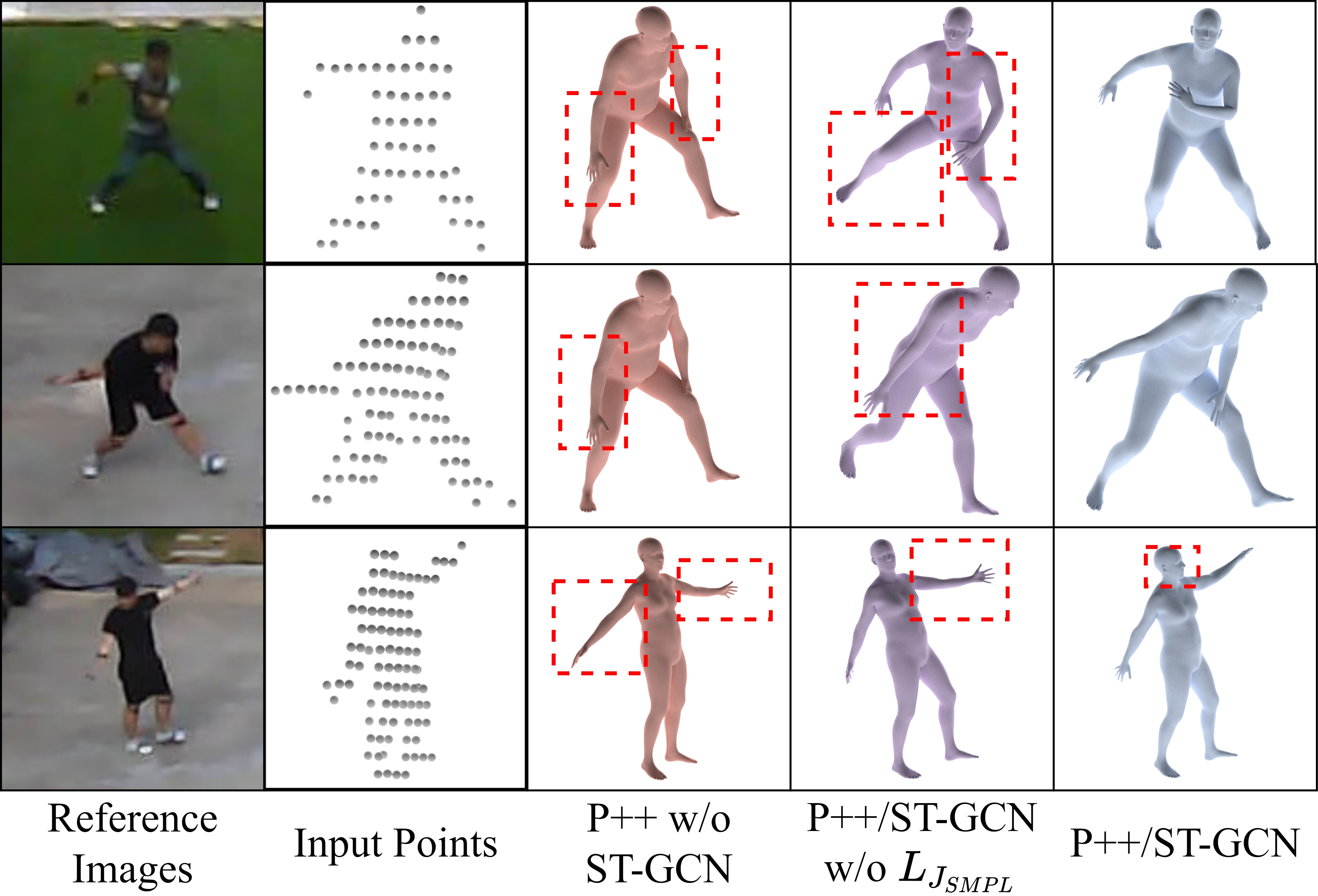}
    \caption{Qualitative results of different combination of stages. Other two methods cannot predict accurate human motions in some cases for lacking enough constraint.}
    \label{fig:stage}
    \vspace{-5mm}
\end{figure}

\subsection{Results on KITTI and Waymo Dataset}

\begin{figure}
  \centering
  \begin{subfigure}{\columnwidth}
    \includegraphics[width=\columnwidth]{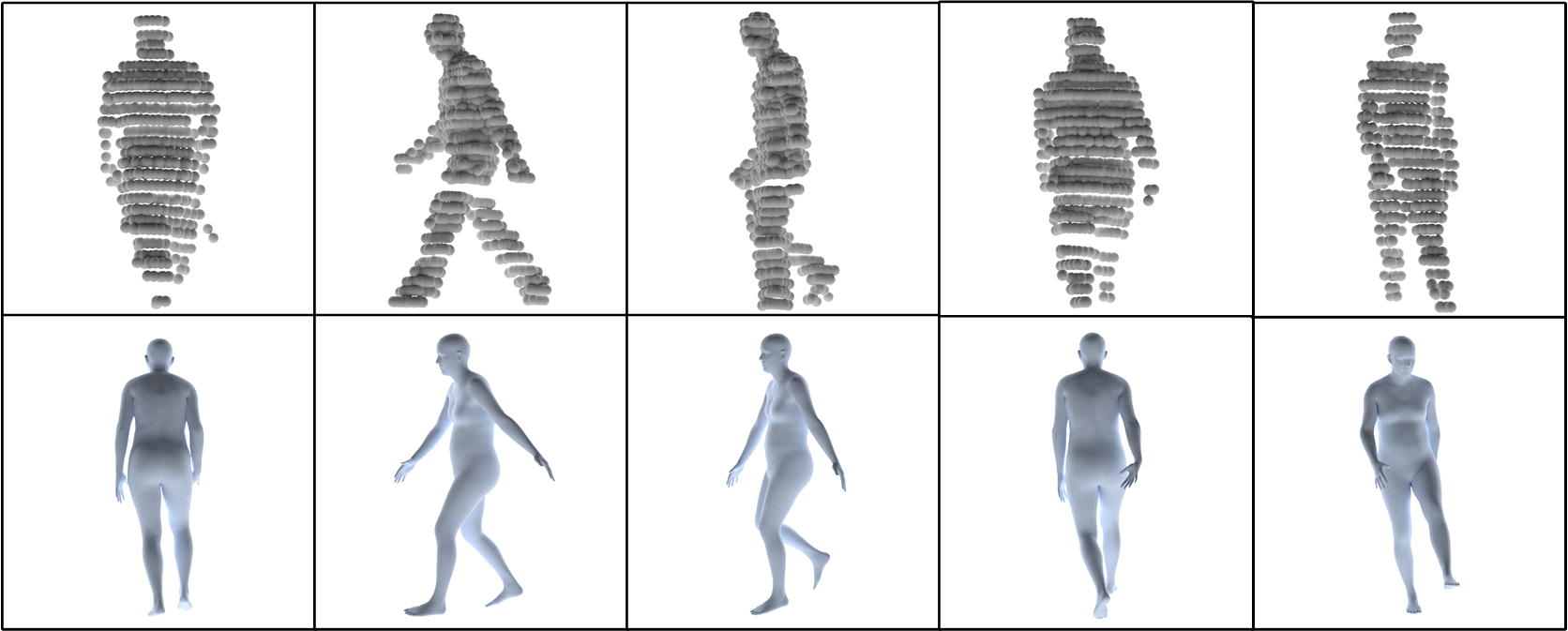}
    \caption{Qualitative results of the KITTI Dataset.}
  \end{subfigure}

  \begin{subfigure}{\columnwidth}
    \includegraphics[width=\columnwidth]{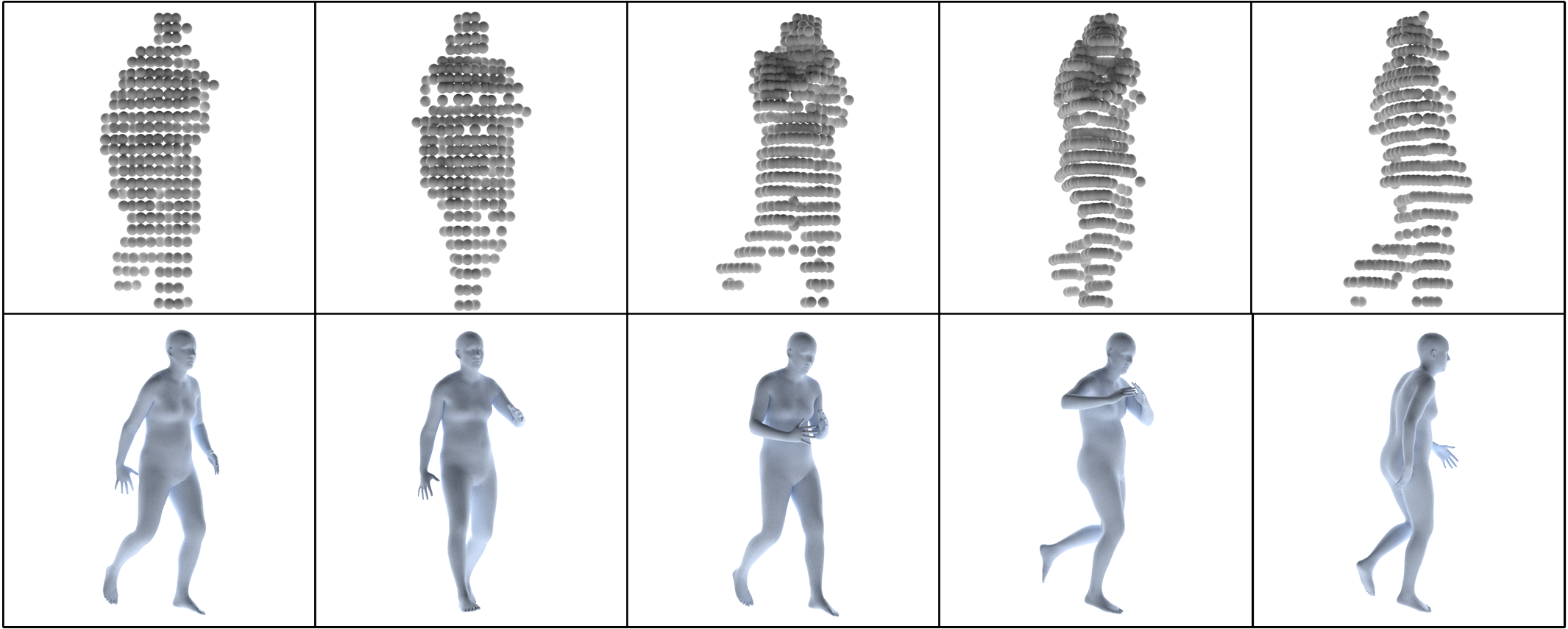}
    \caption{Qualitative results of the Waymo Open Dataset.}
  \end{subfigure}
  
  \begin{subfigure}{\columnwidth}
    \includegraphics[width=\columnwidth]{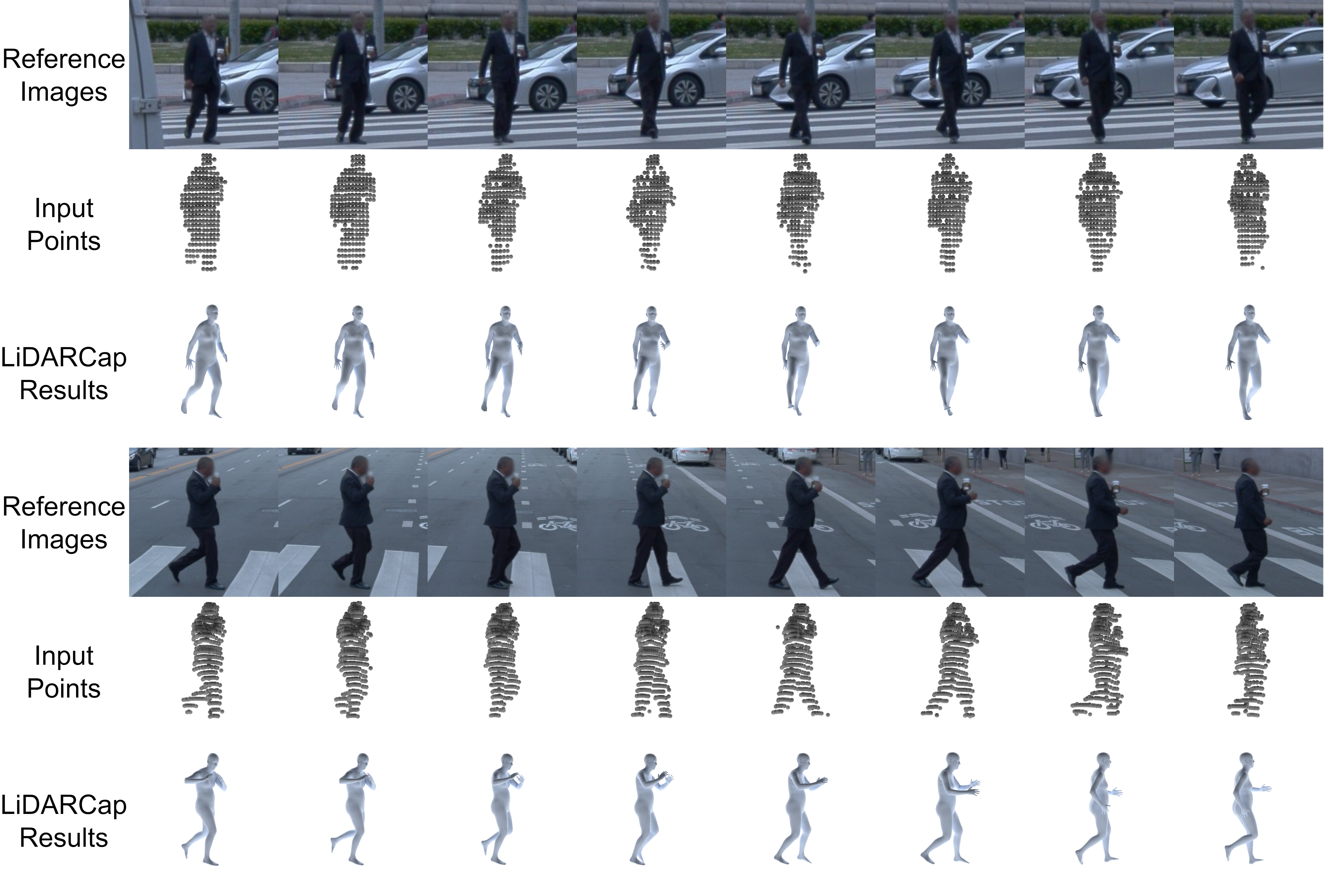}
    \caption{Qualitative results of two sequences from the Waymo Open Dataset.}
  \end{subfigure}
  \vspace{-1mm}
  \caption{Qualitative results on autonomous driving datasets. Our method can discriminate the correct motions of the clear parts of the point cloud and give a reasonable guess of the invisible ones.}
  \vspace{-6mm}
  \label{fig:generalization}
\end{figure}

In order to verify the generalization of our pipeline, we test our method on point cloud sequences of pedestrians from the KITTI Detection Dataset \cite{Geiger2012AreWR} and the Waymo Open Dataset \cite{Sun2020ScalabilityIP}. \cref{fig:generalization} shows some qualitative results.

It can be seen that our algorithm can learn the correct footsteps and global orientation of pedestrians. For the clear part of the upper limb, our method can make the correct placement, while for the ambiguous one, it will make reasonable guesses through prior information of time series.  

%% file: section/discussion.tex
\section{Discussion}
\myparagraph{Limitation.}
First, the scenario in LiDARHuman26M is flat, open, and unobstructed, which is too idealistic compared to the real applications. 
Second, the shape parameters $\boldsymbol{\beta}$ and more complex scenes with occlusions and interactions of multi-person are lacked in the dataset LiDARHuman26M.
Third, the proposed baseline LiDARCap method is not robust enough to 
handle varying density of point clouds from different distances and devices.
Accurate human motion capture on sparse LiDAR point clouds is still an open challenge.

\myparagraph{Conclusion.} We present LiDARHuman26M, the first of its kind dataset to open up the research direction of data-driven LiDAR-based human motion capture in the long-range setting. LiDARHuman26M consists of various modalities, including synchronous LiDAR point clouds, RGB images, and ground-truth  3D  human motions obtained from professional IMU devices. It covers 20 kinds of daily motions and 13 performers with 184.0k capture frames,  with a large capture distance ranging from 12 m to 28 m. Based on LiDARHuman26M, we propose a strong baseline method, LiDARCap, the first marker-less, long-range, and data-driven human motion capture method for monocular LiDAR sensor. Specifically, the proposed LiDARCap extracts global features of LiDAR point clouds. It then employs the inverse kinematics solver and SMPL optimizer to hierarchically regress the human pose through aggregating the temporally encoded features. Quantitative and qualitative experiments show that our method outperforms the methods based only on RGB images. The experiments on the LiDAR data of the KITTI Dataset and the Waymo Open Dataset show that our method can be generalized to different LiDAR sensor settings. 